\documentclass[11pt,a4paper]{article}
\usepackage[hyperref]{acl2021}
\usepackage{latexsym}
\usepackage{times}
\usepackage{booktabs}
\usepackage{graphicx}
\usepackage{multirow}
\usepackage{amssymb}
\usepackage{amsmath}
\usepackage{enumitem}

\usepackage{enumitem}

\usepackage{microtype}

\aclfinalcopy 
\def\aclpaperid{3451} 

\usepackage{todonotes}
\makeatletter
\newcommand*\iftodonotes{\if@todonotes@disabled\expandafter\@secondoftwo\else\expandafter\@firstoftwo\fi} 
\makeatother

\definecolor{RBBlue}{rgb}{0.28,0.25,0.98}
\newcommand\redit[1]{{\color{black} #1}}
\newcommand\dpedit[1]{{\color{black} #1}}

\title{Modulating Language Models with Emotions}

\author{Ruibo Liu$^{1}$ \hspace{0.5cm} Jason Wei$^{2}$ \hspace{0.5cm} Chenyan Jia$^{3}$ \hspace{0.5cm} Soroush Vosoughi$^{1}$\\
  $^{1}$Dartmouth College \hspace{0.5cm} $^{2}$ProtagoLabs \hspace{0.5cm} $^{3}$University of Texas at Austin\\
   \texttt{\href{mailto:ruibo.liu.gr@dartmouth.edu}{ruibo.liu.gr@dartmouth.edu}} \hspace{0.2cm} \texttt{\href{mailto:jason@protagolabs.com}{jason@protagolabs.com}} \\
    \texttt{\href{mailto:chenyanjia@utexas.edu}{chenyanjia@utexas.edu}} \hspace{0.2cm} \texttt{\href{mailto:soroush@dartmouth.edu}{soroush@dartmouth.edu}}  \\}

\date{}

\begin{document}

\maketitle
\begin{abstract}

Generating context-aware language that embodies diverse emotions is an important step towards building empathetic NLP systems. In this paper, we propose a formulation of modulated layer normalization---a technique inspired by computer vision---that allows us to use large-scale language models for emotional response generation. In automatic and human evaluation on the MojiTalk dataset, our proposed modulated layer normalization method outperforms prior baseline methods while maintaining diversity, fluency, and coherence. Our method also obtains competitive performance even when using only 10\% of the available training data.

\end{abstract}

\section{Introduction}

Building interactive systems that can understand and express human emotions has been a long-term goal of artificial intelligence \citep{shen-feng-2020-cdl,huang2018automatic,salovey1997emotional}. Given a context, an intelligent agent ought to be able to generate responses that not only consider the context but also reflect a specified emotion, a task called \textit{emotional response generation}. 
One common representation of emotions is through \textit{emojis}, which often convey the underlying emotions in an utterance \cite{zhou2018mojitalk}. 
Table \ref{tab:exp1_few_shot} shows an example generation in this formulation.

To tackle this problem, prior work has proposed a number of different models, including variants of sequence-to-sequence (Seq2Seq) models~\citep{serban2016building,li2016diversity}, variational autoencoders (VAE)~\citep{gu2019dialogwae,shen-etal-2017-conditional,zhao-etal-2017-learning} and adversarial networks~\citep{kong2019adversarial,li-etal-2017-adversarial}. 
Their generated responses are often dull or generic, partially due to the limited training data for diverse emotions~\citep{li-etal-2017-adversarial}. \dpedit{More recent studies have tried to pre-train language models (LMs) on specific domain data to pivot generation towards certain direction~\cite{gao-etal-2020-dialogue,zhang-etal-2020-dialogpt,yang-etal-2020-styledgpt,keskarCTRL2019}. However, training a LM from scratch can be costly, and collecting sufficient pre-training data in diverse emotions is also challenging, especially for low-resource emotions~\citep{yang-etal-2019-low}.}

\begin{figure}[!t]
\centering
\includegraphics[width=0.50\textwidth]{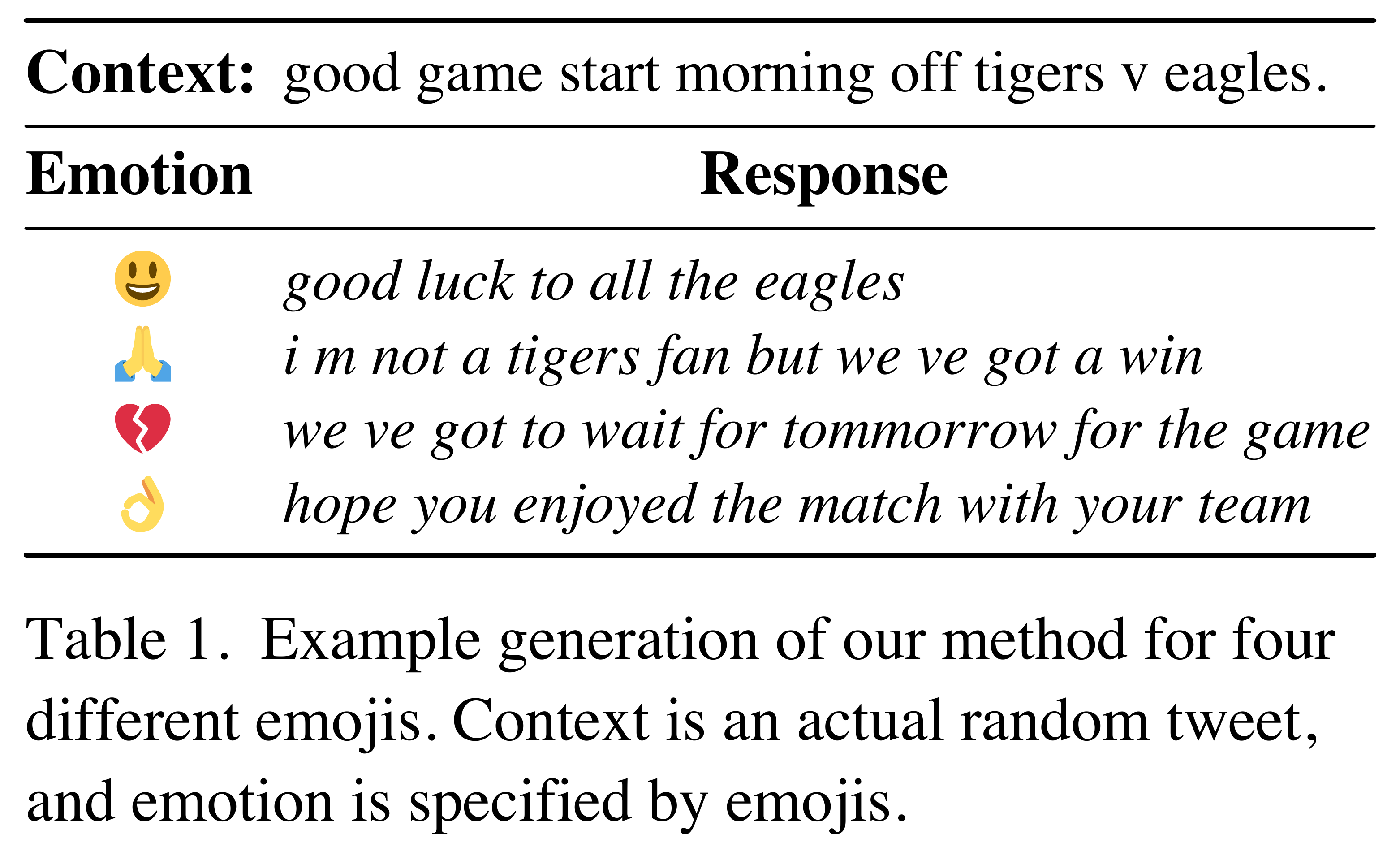}
\end{figure}

\redit{
In this work, we present a simple and easy-to-deploy technique that can enable pre-trained large-scale LMs to generate fine-grained emotional responses. Specifically, we inject emotional signals specified by 64 commonly used emojis via \dpedit{\underline{Mod}ulated \underline{L}ayer \underline{N}ormalization (Mod-LN)}, a technique widely adopted in computer vision but whose potential has not been well studied yet in NLP. The main advantages of our method are:}
\vspace{-0.05in}
\begin{itemize}[leftmargin=\parindent,align=left,labelwidth=\parindent,labelsep=0pt]
     \item Instead of designing or re-training models from scratch, our method is plug-and-play. In this work, we show its effectiveness on BERT~\shortcite{devlin2019bert} and GPT-2~\shortcite{radford2019language}, but one can easily extend our method to other Transformer-based LMs.
     \vspace{-0.05in}
     \item By fully exploiting the transfer learning ability of pre-trained LMs, we achieve comparable emotional response generation performance as prior best-performing work with only 10\% of the training data, which is especially beneficial for low-resource scenarios.
 
 \end{itemize}

\section{Approach}

Given a context text and a specified emoji as a target emotion, we aim to generate responses that both reflect the emotion associated with the emoji and the semantic information in the context. 
In this work, we demonstrate how to inject target emotions through a modulation module of layer normalization ($\S$\ref{sub:cond_ln}). We also provide data preparation and model adaptation strategies on two typical LMs (BERT and GPT-2) to aid reproduction and extension ($\S$\ref{sub:model_adaptation}).

\subsection{Modulated Layer Normalization}
\label{sub:cond_ln}

\begin{figure}[!h]
\centering
\includegraphics[width=0.48\textwidth]{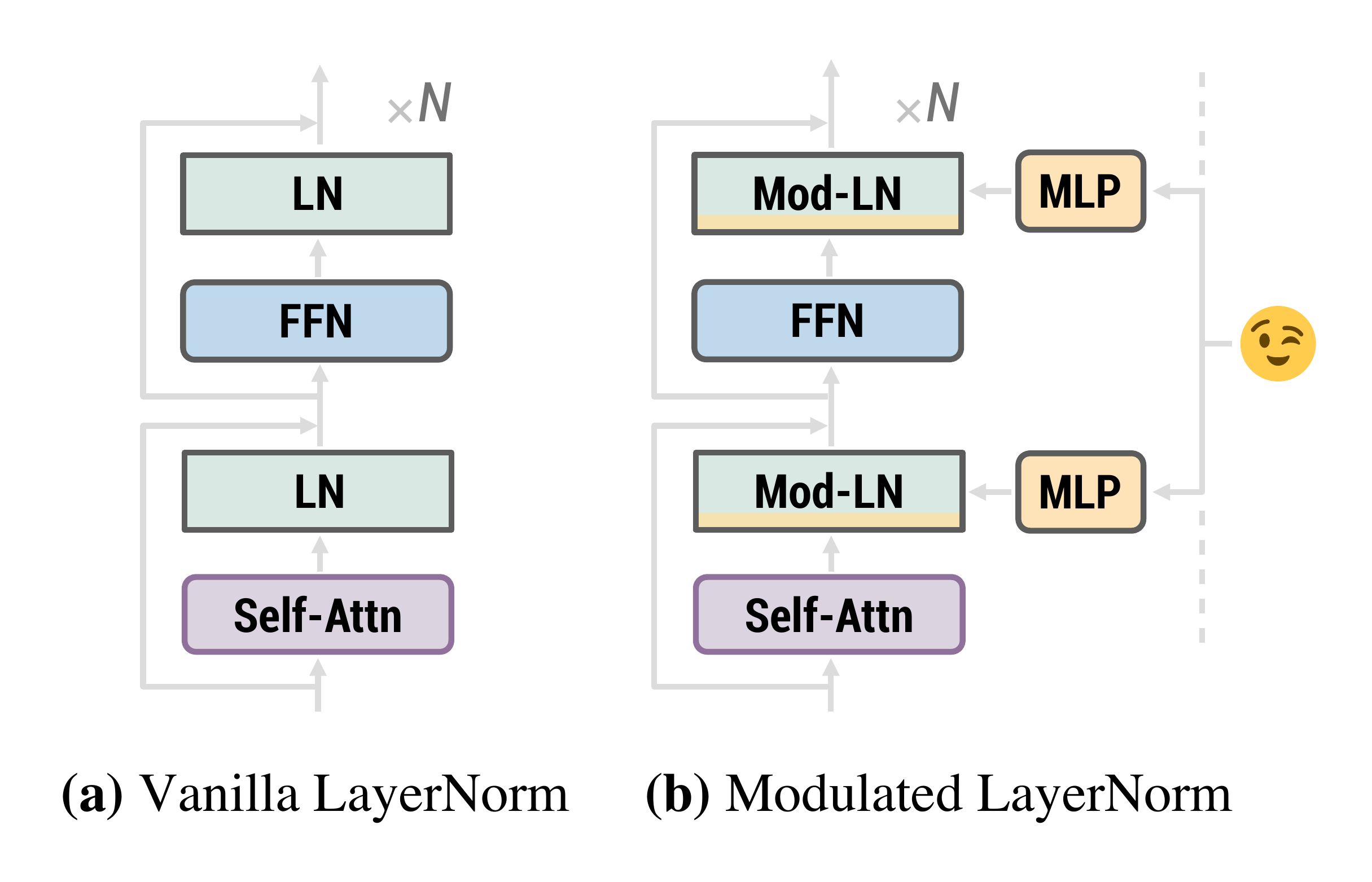}
\caption{Overview of (a) Vanilla Layer Normalization (LN) and (b) Modulated Layer Normalization (Mod-LN) in Transformer-based LMs. \dpedit{The modulation module in Mod-LN uses two Multi-Layer Perceptrons (MLPs) that each have two sets of dense layers. It uses the external emotion input to modulate regularization towards a certain emotion $c$. FFN: Feed-Forward Network. Self-Attn: Multi-head Self-Attention blocks.}}
\label{fig:LN_overview}
\end{figure}

Layerwise-normalization (LN) is commonly used in Transformer-based~\citep{NIPS2017_7181} language models (LMs)~\citep{devlin2019bert,radford2019language,yang2019xlnet} to stabilize hidden state dynamics and reduce training time ~\citep{ba2016layer}. 
In the vanilla implementation (Figure~\ref{fig:LN_overview}(a)), data are normalized by their own mean $\mu$ and standard deviation $\sigma$ \dpedit{without relying on external inputs}.

\dpedit{In contrast to vanilla LN that only regularizes data itself, Mod-LN introduces an external modulation module shared across the whole dataset, which is independent of the individual data samples and able to modulate the regularization towards external inputs $c$ (Figure~\ref{fig:LN_overview} (b))}. Specifically, for an input hidden state tensor $x$ in layer $l$, it is normalized by Mod-LN as
\begin{equation}
\label{eqa:cond_norm}
    x = \textrm{MLP}^{(l)}_{\gamma}(c) \cdot \frac{x - \mu}{\sigma + \epsilon} + \textrm{MLP}^{(l)}_{\beta} \ ,
\end{equation}
where $\epsilon$ is the smoothing parameter to avoid dividing by zero. \dpedit{$\textrm{MLP}^{(l)}_{\gamma}$ and $\textrm{MLP}^{(l)}_{\beta}$ are two trainable modulation modules for a certain layer $l$. They are computed by }

\begin{align}
    \textrm{MLP}^{(l)}_{\gamma}(c) &= W_{\gamma}^{(l, 2)} \cdot \textrm{Swish} (W_{\gamma}^{(l, 1)} c), \\
    \textrm{MLP}^{(l)}_{\beta}(c) &= W_{\beta}^{(l, 2)} \cdot \textrm{Swish} (W_{\beta}^{(l, 1)} c + b),
\end{align}
where $W^{(l, 1)}$ and $W^{(l, 2)}$ are dense layers belonging to layer $l$, with weights size of [64, $\frac{1}{2} \cdot \textrm{dim}_h$] and [$\frac{1}{2} \cdot \textrm{dim}_h$, $\textrm{dim}_h$] respectively\footnote{The hidden size $\textrm{dim}_h$ of bert-large-uncased and GPT-2 medium model are both 1024.}. Dense layers connect 64 emoji classes to the output hidden states of the language model, and $b$ is a bias added to $\gamma$. 
We use the Swish activation~\citep{ramachandran2017searching}, which has been shown to outperform  ReLU~\citep{xu2015empirical} on several challenging datasets. \dpedit{Though conceptually simple, such MLP based modules have been shown to be a faster and more efficient alternative to vanilla dot product self-attention in NLP~\citep{tay2020synthesizer} and CV~\cite{tolstikhin2021mlp}. Our work uses MLPs as a plug-and-play modulator rather than a replacement for self-attentions, allowing us to shift the hidden states towards a given target emotion.} 

\subsection{Data Preparation and Model Adaptation}
\label{sub:model_adaptation}

For the text input, we concatenate ground-truth context with corresponding response as a whole input to feed into LMs. We add a pre-defined separator token (\texttt{[SEP]} for BERT and \texttt{[UNK]} for GPT-2) between context and response, to make LMs aware of the range of each part. \dpedit{We also pad both context and response to a max sequence length with the padding token.}

\dpedit{Encoder-Decoder models have been successful in many text-to-text generation tasks, such as question answering~\citep{chen2017reading,seo2016bidirectional}, news summarization~\citep{chopra-etal-2016-abstractive,rush2015neural}, and style transfer~\citep{li-etal-2018-delete,liu2021transformer}. For the response generation task, the encoder encodes the context text into a fixed-length vector in latent space, while the decoder decodes the generated response tokens step-by-step, given the encoded context vector and the ground truth token from the previous step; this method is also known as teacher-forcing~\citep{zhang-etal-2019-bridging,cho-etal-2014-learning}.

In this work, we consider leveraging the transfer learning power of large-scale LMs---using LMs as encoder and decoder---to better capture the complicated relationship between context and response~\citep{rothe2020leveraging}. Auto-regressive LMs (ARLMs), such as GPT-2 are trained to iteratively predict the next step token given the past, while Masked Language Models (MLM), such as BERT, are trained to predict missing tokens given both the preceding and subsequent text. In contrast to the uni-directional attention flow in ARLM, the attention flow of MLM is bi-directional, and thus if we directly use MLM as decoder, the prediction of tokens in the response will also attend to (i.e., have the context of) future tokens; this could potentially lead to exposure bias~\cite{schmidt-2019-generalization}. Inspired by recent text-to-text LMs such as T5~\citep{raffel2020exploring} and BART~\citep{lewis2019bart}, for MLM decoder, we
modify the original bi-directional attention mask to make it uni-directional. 

We experiment with two encoder-decoder models built on MLM and ARLM: 1) BERT-to-BERT: using bi-directional BERT as both encoder and decoder, but forcing the decoder BERT to attend to past context with uni-directional mask, and 2) GPT2-to-GPT2: using uni-directional GPT-2 as both encoder and decoder.
}

\section{Experimental Setup}

\noindent \textbf{Dataset.} For all the experiments, we use the MojiTalk~\citep{zhou2018mojitalk} dataset, a large Twitter conversation corpus ($N\approx700k$) of responses that each contain one or more of 64 popular emojis. Following the original paper, we split the corpus into training, validation, and test sets of 596,959, 32,600, and 32,600 conversation pairs, respectively. We fine-tune \dpedit{the two LM-based encoder-decoder models on this dataset and generate responses given contexts and all possible emotions using top-$k$ random decoding \citep{fan2018hierarchical} on 
a machine with four RTX 2080 GPUs}~\footnote{We choose $k=10$ for a balance of generation diversity and readability through empirical observation.}.

\vspace{1.5mm} \noindent \textbf{Models.} We evaluate three models in total. 
We take the \underline{R}einforced \underline{C}onditional \underline{V}ariational \underline{A}uto\underline{E}ncoders (R-CVAE) model from \citet{zhou2018mojitalk} as \textbf{Baseline} (current best-performing model on 64-emoji controlled response generation), \textbf{Mod-LN MLM}: BERT-to-BERT (large, uncased) + Mod-LN, and \textbf{Mod-LN ARLM}: GPT2-to-GPT2 (large) + Mod-LN.

\begingroup
\setlength{\tabcolsep}{2pt}
\begin{table}[!ht]
\centering
\resizebox{0.41\textwidth}{!}{%
\begin{tabular}{@{}lccc@{}}
\toprule
\multicolumn{1}{c}{\multirow{2}{*}{\textbf{Model}}} & \multicolumn{3}{c}{\textbf{Emoji Acc (\%)}} \\ \cmidrule(l){2-4} 
\multicolumn{1}{c}{}           & Hits@1                & Hits@3                & Hits@5                \\ \midrule
\textbf{Baseline}: R-CVAE                         &                      &                      &                      \\
w/. 10\% train data  & 13.4                 & 27.1                 & 33.6                 \\
w/. 100\% train data & 26.2                 & 44.2                 & 53.4                 \\ \midrule
\textbf{Mod-LN MLM}            & \multicolumn{1}{l}{} & \multicolumn{1}{l}{} & \multicolumn{1}{l}{} \\
w/. 10\% train data  & 20.5                 & \underline{47.4}                 & 59.1                 \\
w/. 100\% train data & 33.6                 & 56.8                 & 72.2                 \\ \midrule
\textbf{Mod-LN ARLM}             & \multicolumn{1}{l}{} & \multicolumn{1}{l}{} & \multicolumn{1}{l}{} \\
w/. 10\% train data  & \underline{27.9}                 & 43.4                 & \underline{64.1}                 \\
w/. 100\% train data & \underline{34.4}                 & \underline{60.3}                 & \underline{82.5}                 \\ \bottomrule
\end{tabular}%
}
\caption{Accuracy of emotional response judged by DeepMoji on classifying emotions in responses generated by R-CVAE, MLM (BERT) with Mod-LN, and ARLM (GPT-2) with Mod-LN.}
\label{tab:exp1_few_shot}
\end{table}
\endgroup

\section{Evaluation} 
Good emotional responses should accurately reflect the intended emotion, be diverse, and have coherent language. We thus evaluate three aspects of generated responses: emotion control ($\S$\ref{subsec:emotion_control}), response diversity ($\S$\ref{subsec:gen_divers}), and coherence and fluency ($\S$\ref{subsec:coherence_fluency}). 
We also use Amazon Mechanical Turk (MTurk) to run a manual evaluation of emotion control and readability in generated responses ($\S$\ref{subsec:human_evaluation}).

\subsection{Emotion Control}
\label{subsec:emotion_control}

First, we evaluate whether intended emotions were reflected in the responses generated by various models. We choose DeepMoji~\citep{felbo-etal-2017-using}\footnote{We chose the official implementation by huggingface: https://github.com/huggingface/torchMoji.} as the judgment classifier. DeepMoji was trained on a large-scale emoji dataset containing 1,246 million tweets and 64 distinct emojis, and as far as we know, is state-of-the-art for 64-emoji classification tasks. Since the meanings of different emojis can overlap with subtle differences, we compute Hits@$k$ ($k$ = \{1, 3, 5\}) classification accuracy~\cite{gao-etal-2020-dialogue} to describe the performance of models in different criteria.
As shown in Table \ref{tab:exp1_few_shot}, our proposed models outperform R-CVAE with a large margin. Of note, LM-based models reveal more robust performance in extreme data scarcity cases: our models achieve comparable performance with R-CVAE even when using only 10\% of the training data. Between BERT and GPT-2, GPT-2 shows superior performance, partially because its weights are from auto-regressive pre-training.


\begingroup
\setlength{\tabcolsep}{2pt}
\begin{table}[!t]
\centering
\resizebox{0.48\textwidth}{!}{%
\begin{tabular}{@{}lcccc@{}}
\toprule
\multicolumn{1}{c}{\textbf{Model}} & \textbf{TTR-1} & \textbf{TTR-2} & \multicolumn{1}{l}{\textbf{Avg. len}} & \multicolumn{1}{l}{\textbf{\%stop}} \\ \midrule
Human Reference & 0.059 & 0.43 & 11.7 & 50.4 \\ \midrule
\textbf{Baseline}: R-CVAE               &       &      &      &      \\
w/. 10\% train data   & 0.034 & 0.24 & 8.6  & 60.1 \\
w/. 100\% train data  & 0.051 & 0.33 & 9.2  & 59.3 \\ \midrule
\textbf{Mod-LN MLM}   &       &      &      &      \\
w/. 10\% train data   & 0.054 & \underline{0.43} & 18.2 & 49.3 \\
w/. 100\% train data  & \underline{0.059} & 0.39 & 14.3 & 49.2 \\ \midrule
\textbf{Mod-LN ARLM}  &       &      &      &      \\
w/. 10\% train data   & 0.056 & 0.38 & 15.9 & 48.7 \\
w/. 100\% train data  & 0.057 & 0.40 & 12.5 & 48.5 \\ \bottomrule
\end{tabular}%
}
\caption{Lexical diversity of generated responses from various models. 
TTR-1/TTR-2: unigram/bigram type-token ratio;
Avg. len: average number of tokens in generated responses; 
\%stop: average percent of stop words among all tokens in the generated responses. }
\label{tab:diversity}
\end{table}
\endgroup

\subsection{Generation Diversity}
\label{subsec:gen_divers}

As shown in Table \ref{tab:diversity}, we evaluate the diversity of responses generated by each model in terms of unigram and bigram type-token ratios, average length, and percent of stop words in generated responses, with values for the human-generated responses shown for reference.
As measured by the type-token ratio for both uni- and bi-grams, our proposed models generate more diverse responses. 
In addition, compared with the R-CVAE, the responses generated by our models are longer and use fewer stop words. The advance can be attributed to the using of large-scale language models as base models.

\subsection{Fluency and Coherence}
\label{subsec:coherence_fluency}
Moreover, we evaluate the fluency and coherence of machine-generated text. 
For fluency, we trained a standalone language model on the human-generated responses using KenLM~\citep{heafield2011kenlm} to measure the perplexity of generated texts. 
To evaluate coherence between the context and the generated responses, we compute the similarity between the generated text and human-generated responses using BERTScore~\citep{bertscore}, with the human-generated responses as reference. We configure the BERTScore using 24-layer RoBERTa-large~\citep{liu2019roberta} as for English tasks.
Table~\ref{tab:coherence} shows these results.
For perplexity and BERTScore, our Mod-LN models outperform the R-CVAE in both 10\% and 100\% training data cases.

\begingroup
\setlength{\tabcolsep}{2pt}
\begin{table}[!t]
\centering
\resizebox{0.44\textwidth}{!}{%
\begin{tabular}{@{}lcccc@{}}
\toprule
\multirow{2}{*}{\textbf{Model (vs Ref.)}} & \multirow{2}{*}{\textbf{PPL}} & \multicolumn{3}{c}{\textbf{BERTScore (\%)}} \\ \cmidrule(l){3-5} 
                    &        & Precision & Recall & F1   \\ \midrule
\textbf{Baseline}: R-CVAE              &        &           &        &      \\
w/. 10\% train data  & 121.18 & 74.9      & 83.0   & 76.7 \\
w/. 100\% train data & 92.64  & 80.8      & 80.8   & 80.8 \\ \midrule
\textbf{Mod-LN MLM}  &        &           &        &      \\
w/. 10\% train data  & 79.24  & 78.4      & 80.1   & 78.8 \\
w/. 100\% train data & 50.72  & 82.9      & 84.1   & 83.5 \\ \midrule
\textbf{Mod-LN ARLM} &        &           &        &      \\
w/. 10\% train data  & 51.55  & 83.7      & 80.7   & 83.2 \\
w/. 100\% train data & \underline{36.31}  & 84.7      & 86.2   & \underline{85.4} \\ \bottomrule
\end{tabular}%
}
\caption{Fluency as measured by perplexity (PPL) and coherence as measured by BERTScore of generated responses from various models. Ref.: Human-generated responses.}
\label{tab:coherence}
\end{table}
\endgroup

\subsection{Human Evaluation}
\label{subsec:human_evaluation}

\dpedit{
In total 120 MTurk participants manually evaluated the emotion control and readability of responses from our proposed models and the original human-generated reference data. The average age of participants was 38.40 years-old (SD = 12.26, Median=34.50). More than half (65.8\%) of participants were male, and 34.2\% were female. The average completion time of each survey was 4.53 minutes. Participants were paid \$1 per survey, averaging to more than \$13 per hour wage for each participant, significantly above the U.S. federal minimum wage.

\paragraph{Procedure} Each participant was assigned to read five randomly selected context-response pairs without being informed of the sources of the responses. They were asked to rate 1) emotion control: ``\textit{How well
the emotion conveyed in the response agrees with
the specified emoji?} (1-very well to 7-not at all)'', and 2) readability: ``\textit{Please rate the readability of the response on a 7-point scale.} (1-very low to 7-very high)''. The readability measure included five items adapted from a previous study~\citep{graefe2018readers}, specifically, well-written, concise, comprehensive, coherent, and clear. Since the five measures had very high agreement (Cronbach’s\footnote{Cronbach's alpha is a measure of internal consistency between sets of items.} $\alpha$ = .91), we average the five measures into one as a general readability index.

\paragraph{Results} The participant's averaged ratings ($\mu$) and Standard Errors (SE) are reported in Table~\ref{tab:human_judgement}. As shown in the table, the standard error of the mean among all annotators is .10, which is very low for a 7-point scale, indicating large agreement between annotators. Responses generated by Mod-LN MLM (BERT), Mod-LN ARLM (GPT-2), and the human-generated references had no statistically significant differences in emotion control and readability. All were rated significantly higher than plain GPT-2 and R-CVAE in both emotion control and readability ($p <$ .001 for one-way repeated measures ANOVA). We also conducted pairwise multiple comparisons in our analysis as post-hoc analysis. In terms of emotion control, both of our two proposed models and original reference data were rated significantly better than vanilla GPT-2 ($p <$ .007). For readability, both our models, vanilla GPT-2, and original reference data were rated significantly more readable than R-CVAE ($p <$ .001).
}

\begingroup
\setlength{\tabcolsep}{2pt}
\begin{table}[]
\centering
\resizebox{0.48\textwidth}{!}{%
\begin{tabular}{@{}lcc@{}}
\toprule
\multirow{2}{*}{\textbf{Response Source}} & \multicolumn{2}{c}{\textbf{Annotator Ratings}: $\mu$ (SE)} \\ \cmidrule(l){2-3} 
                                 & \textsc{Emo Ctrl.}      & \textsc{Readability}     \\ \midrule
Human Reference                           & 5.62 (0.10)           & 5.34 (0.10)      \\ \midrule
\textbf{Baseline}: R-CVAE                        & 4.80 (0.10)           & 4.67 (0.10)      \\ \midrule
\textbf{Mod-LN MLM}                       & 5.43 (0.10)           & 5.20 (0.10)      \\ \midrule
\textbf{Ablation}: Vanilla GPT-2                            & 4.98 (0.10)           & 4.64 (0.10)      \\
\textbf{Mod-LN ARLM}            & 5.40 (0.10)           & 5.32 (0.10)      \\ \bottomrule
\end{tabular}%
}
\caption{Humans manually evaluated the emotional control and readability of responses from the original data (human reference), Baseline and proposed models on a 7-point scale (1: low quality, 7: high quality). We also take the generative LM: vanilla GPT-2, as the ablation reference.}
\label{tab:human_judgement}
\end{table}
\endgroup

\section{Related Work}
\label{sec:related}

\dpedit{
\textbf{Emotional Text Generation.}} VAE-based models \citep{park2018hierarchical,shen-etal-2017-conditional,zhao-etal-2017-learning,serban2017hierarchical}, adversarial networks \citep{kong2019adversarial,li-etal-2017-adversarial,yu2017seqgan} and reinforcement learning systems \citep{li2019dialogue,li-etal-2016-deep} have dominated sentiment-aware dialogue models. \dpedit{Other methods have been developed using LSTM~\citep{song-etal-2019-generating} and GRU~\citep{wei2019emotion,zhou2018emotional}.
All these methods, however, are built on relatively coarse emotion types, partially due to the limited modeling ability of RNNs. Our model outperforms current state-of-the-art R-CVAE~\citep{zhou2018mojitalk} in the same 64-emoji settings.}

\vspace{1mm} \noindent \dpedit{\textbf{Modulated Normalization.}} Though not common in NLP, modulated normalization has been previously used in computer vision. 
In addition to work mentioned in the introduction \citep{de2017modulating}, adversarial networks such as CGAN~\citep{miyato2018cgans}, self-attention GAN~\citep{zhang2019self} and Style GAN~\citep{karras2019style} have used modulated normalization to inject external signal into their models. \dpedit{In NLP, previous studies have tried to modulate normalization for classification tasks~\citep{pmlr-v97-houlsby19a} and multilingual machine translation~\citep{bapna-firat-2019-simple}, however, both these methods require architecture-level modifications. Our method, on the other hand, is plug-and-play, requiring minimal modifications to the architecture and thus easier to deploy for a diverse set of applications.}

\section{Conclusions}

We have proposed a modulated layer normalization approach to generating responses of varying specified emotions. Our approach allows us to leverage large pre-trained models, while remaining simple and easily-extendable. In empirical experiments, our approach substantially outperforms prior work and achieves comparable results using only 10\% of the available training data, all while maintaining diversity, fluency, and coherence.

\bibliographystyle{acl_natbib}
\bibliography{acl2021}

\begin{thebibliography}{57}
\expandafter\ifx\csname natexlab\endcsname\relax\def\natexlab#1{#1}\fi

\bibitem[{Ba et~al.(2016)Ba, Kiros, and Hinton}]{ba2016layer}
Jimmy~Lei Ba, Jamie~Ryan Kiros, and Geoffrey~E Hinton. 2016.
\newblock \href {https://arxiv.org/pdf/1607.06450.pdf} {Layer normalization}.
\newblock In \emph{Advances in Neural Information Processing Systems, Deep
  Learning Symposium}.

\bibitem[{Bapna and Firat(2019)}]{bapna-firat-2019-simple}
Ankur Bapna and Orhan Firat. 2019.
\newblock \href {https://doi.org/10.18653/v1/D19-1165} {Simple, scalable
  adaptation for neural machine translation}.
\newblock In \emph{Proceedings of the 2019 Conference on Empirical Methods in
  Natural Language Processing and the 9th International Joint Conference on
  Natural Language Processing (EMNLP-IJCNLP)}, pages 1538--1548, Hong Kong,
  China. Association for Computational Linguistics.

\bibitem[{Chen et~al.(2017)Chen, Fisch, Weston, and Bordes}]{chen2017reading}
Danqi Chen, Adam Fisch, Jason Weston, and Antoine Bordes. 2017.
\newblock \href {https://www.aclweb.org/anthology/P17-1171.pdf} {Reading
  wikipedia to answer open-domain questions}.
\newblock In \emph{Proceedings of the 55th Annual Meeting of the Association
  for Computational Linguistics (Volume 1: Long Papers)}, pages 1870--1879.

\bibitem[{Cho et~al.(2014)Cho, van Merri{\"e}nboer, Gulcehre, Bahdanau,
  Bougares, Schwenk, and Bengio}]{cho-etal-2014-learning}
Kyunghyun Cho, Bart van Merri{\"e}nboer, Caglar Gulcehre, Dzmitry Bahdanau,
  Fethi Bougares, Holger Schwenk, and Yoshua Bengio. 2014.
\newblock \href {https://doi.org/10.3115/v1/D14-1179} {Learning phrase
  representations using {RNN} encoder{--}decoder for statistical machine
  translation}.
\newblock In \emph{Proceedings of the 2014 Conference on Empirical Methods in
  Natural Language Processing ({EMNLP})}, pages 1724--1734, Doha, Qatar.
  Association for Computational Linguistics.

\bibitem[{Chopra et~al.(2016)Chopra, Auli, and
  Rush}]{chopra-etal-2016-abstractive}
Sumit Chopra, Michael Auli, and Alexander~M. Rush. 2016.
\newblock \href {https://doi.org/10.18653/v1/N16-1012} {Abstractive sentence
  summarization with attentive recurrent neural networks}.
\newblock In \emph{Proceedings of the 2016 Conference of the North {A}merican
  Chapter of the Association for Computational Linguistics: Human Language
  Technologies}, pages 93--98, San Diego, California. Association for
  Computational Linguistics.

\bibitem[{De~Vries et~al.(2017)De~Vries, Strub, Mary, Larochelle, Pietquin, and
  Courville}]{de2017modulating}
Harm De~Vries, Florian Strub, J{\'e}r{\'e}mie Mary, Hugo Larochelle, Olivier
  Pietquin, and Aaron~C Courville. 2017.
\newblock \href
  {http://papers.nips.cc/paper/7237-modulating-early-visual-processing-by-language.pdf}
  {Modulating early visual processing by language}.
\newblock In \emph{Advances in Neural Information Processing Systems}, pages
  6594--6604.

\bibitem[{Devlin et~al.(2019)Devlin, Chang, Lee, and
  Toutanova}]{devlin2019bert}
Jacob Devlin, Ming-Wei Chang, Kenton Lee, and Kristina Toutanova. 2019.
\newblock \href {https://arxiv.org/pdf/1810.04805.pdf} {Bert: Pre-training of
  deep bidirectional transformers for language understanding}.
\newblock In \emph{Proceedings of the 2019 Conference of the North American
  Chapter of the Association for Computational Linguistics: Human Language
  Technologies, Volume 1 (Long and Short Papers)}, pages 4171--4186.

\bibitem[{Fan et~al.(2018)Fan, Lewis, and Dauphin}]{fan2018hierarchical}
Angela Fan, Mike Lewis, and Yann Dauphin. 2018.
\newblock \href {https://arxiv.org/pdf/1805.04833.pdf} {Hierarchical neural
  story generation}.
\newblock In \emph{Proceedings of the 56th Annual Meeting of the Association
  for Computational Linguistics (Volume 1: Long Papers)}, pages 889--898.

\bibitem[{Felbo et~al.(2017)Felbo, Mislove, S{\o}gaard, Rahwan, and
  Lehmann}]{felbo-etal-2017-using}
Bjarke Felbo, Alan Mislove, Anders S{\o}gaard, Iyad Rahwan, and Sune Lehmann.
  2017.
\newblock \href {https://doi.org/10.18653/v1/D17-1169} {Using millions of emoji
  occurrences to learn any-domain representations for detecting sentiment,
  emotion and sarcasm}.
\newblock In \emph{Proceedings of the 2017 Conference on Empirical Methods in
  Natural Language Processing}, pages 1615--1625, Copenhagen, Denmark.
  Association for Computational Linguistics.

\bibitem[{Gao et~al.(2020)Gao, Zhang, Galley, Brockett, and
  Dolan}]{gao-etal-2020-dialogue}
Xiang Gao, Yizhe Zhang, Michel Galley, Chris Brockett, and Bill Dolan. 2020.
\newblock \href {https://doi.org/10.18653/v1/2020.emnlp-main.28} {Dialogue
  response ranking training with large-scale human feedback data}.
\newblock In \emph{Proceedings of the 2020 Conference on Empirical Methods in
  Natural Language Processing (EMNLP)}, pages 386--395, Online. Association for
  Computational Linguistics.

\bibitem[{Graefe et~al.(2018)Graefe, Haim, Haarmann, and
  Brosius}]{graefe2018readers}
Andreas Graefe, Mario Haim, Bastian Haarmann, and Hans-Bernd Brosius. 2018.
\newblock Readers’ perception of computer-generated news: Credibility,
  expertise, and readability.
\newblock \emph{Journalism}, 19(5):595--610.

\bibitem[{Gu et~al.(2019)Gu, Cho, Ha, and Kim}]{gu2019dialogwae}
Xiaodong Gu, Kyunghyun Cho, Jung~Woo Ha, and Sunghun Kim. 2019.
\newblock \href {https://arxiv.org/pdf/1805.12352.pdf} {Dialogwae: Multimodal
  response generation with conditional wasserstein auto-encoder}.
\newblock In \emph{7th International Conference on Learning Representations,
  ICLR 2019}.

\bibitem[{Heafield(2011)}]{heafield2011kenlm}
Kenneth Heafield. 2011.
\newblock \href {https://www.aclweb.org/anthology/W11-2123.pdf} {Kenlm: Faster
  and smaller language model queries}.
\newblock In \emph{Proceedings of the sixth workshop on statistical machine
  translation}, pages 187--197. Association for Computational Linguistics.

\bibitem[{Houlsby et~al.(2019)Houlsby, Giurgiu, Jastrzebski, Morrone,
  De~Laroussilhe, Gesmundo, Attariyan, and Gelly}]{pmlr-v97-houlsby19a}
Neil Houlsby, Andrei Giurgiu, Stanislaw Jastrzebski, Bruna Morrone, Quentin
  De~Laroussilhe, Andrea Gesmundo, Mona Attariyan, and Sylvain Gelly. 2019.
\newblock \href {http://proceedings.mlr.press/v97/houlsby19a.html}
  {Parameter-efficient transfer learning for {NLP}}.
\newblock In \emph{Proceedings of the 36th International Conference on Machine
  Learning}, volume~97 of \emph{Proceedings of Machine Learning Research},
  pages 2790--2799. PMLR.

\bibitem[{Huang et~al.(2018)Huang, Zaiane, Trabelsi, and
  Dziri}]{huang2018automatic}
Chenyang Huang, Osmar~R Zaiane, Amine Trabelsi, and Nouha Dziri. 2018.
\newblock \href {https://www.aclweb.org/anthology/N18-2008.pdf} {Automatic
  dialogue generation with expressed emotions}.
\newblock In \emph{Proceedings of the 2018 Conference of the North American
  Chapter of the Association for Computational Linguistics: Human Language
  Technologies, Volume 2 (Short Papers)}, pages 49--54.

\bibitem[{Karras et~al.(2019)Karras, Laine, and Aila}]{karras2019style}
Tero Karras, Samuli Laine, and Timo Aila. 2019.
\newblock \href
  {http://openaccess.thecvf.com/content_CVPR_2019/papers/Karras_A_Style-Based_Generator_Architecture_for_Generative_Adversarial_Networks_CVPR_2019_paper.pdf}
  {A style-based generator architecture for generative adversarial networks}.
\newblock In \emph{Proceedings of the IEEE conference on computer vision and
  pattern recognition}, pages 4401--4410.

\bibitem[{Keskar et~al.(2019)Keskar, McCann, Varshney, Xiong, and
  Socher}]{keskarCTRL2019}
Nitish~Shirish Keskar, Bryan McCann, Lav Varshney, Caiming Xiong, and Richard
  Socher. 2019.
\newblock {CTRL - A Conditional Transformer Language Model for Controllable
  Generation}.
\newblock \emph{arXiv preprint arXiv:1909.05858}.

\bibitem[{Kong et~al.(2019)Kong, Li, Neubig, Hovy, and
  Yang}]{kong2019adversarial}
Xiang Kong, Bohan Li, Graham Neubig, Eduard Hovy, and Yiming Yang. 2019.
\newblock \href {https://arxiv.org/pdf/1901.07129.pdf} {An adversarial approach
  to high-quality, sentiment-controlled neural dialogue generation}.
\newblock \emph{arXiv preprint arXiv:1901.07129}.

\bibitem[{Lewis et~al.(2019)Lewis, Liu, Goyal, Ghazvininejad, Mohamed, Levy,
  Stoyanov, and Zettlemoyer}]{lewis2019bart}
Mike Lewis, Yinhan Liu, Naman Goyal, Marjan Ghazvininejad, Abdelrahman Mohamed,
  Omer Levy, Ves Stoyanov, and Luke Zettlemoyer. 2019.
\newblock \href {https://arxiv.org/pdf/1910.13461.pdf} {Bart: Denoising
  sequence-to-sequence pre-training for natural language generation,
  translation, and comprehension}.
\newblock \emph{arXiv preprint arXiv:1910.13461}.

\bibitem[{Li et~al.(2016{\natexlab{a}})Li, Galley, Brockett, Gao, and
  Dolan}]{li2016diversity}
Jiwei Li, Michel Galley, Chris Brockett, Jianfeng Gao, and Bill Dolan.
  2016{\natexlab{a}}.
\newblock \href {https://www.aclweb.org/anthology/N16-1014} {A
  diversity-promoting objective function for neural conversation models}.
\newblock In \emph{Proceedings of NAACL-HLT}, pages 110--119.

\bibitem[{Li et~al.(2016{\natexlab{b}})Li, Monroe, Ritter, Jurafsky, Galley,
  and Gao}]{li-etal-2016-deep}
Jiwei Li, Will Monroe, Alan Ritter, Dan Jurafsky, Michel Galley, and Jianfeng
  Gao. 2016{\natexlab{b}}.
\newblock \href {https://doi.org/10.18653/v1/D16-1127} {Deep reinforcement
  learning for dialogue generation}.
\newblock In \emph{Proceedings of the 2016 Conference on Empirical Methods in
  Natural Language Processing}, pages 1192--1202, Austin, Texas. Association
  for Computational Linguistics.

\bibitem[{Li et~al.(2017)Li, Monroe, Shi, Jean, Ritter, and
  Jurafsky}]{li-etal-2017-adversarial}
Jiwei Li, Will Monroe, Tianlin Shi, S{\'e}bastien Jean, Alan Ritter, and Dan
  Jurafsky. 2017.
\newblock \href {https://doi.org/10.18653/v1/D17-1230} {Adversarial learning
  for neural dialogue generation}.
\newblock pages 2157--2169.

\bibitem[{Li et~al.(2018)Li, Jia, He, and Liang}]{li-etal-2018-delete}
Juncen Li, Robin Jia, He~He, and Percy Liang. 2018.
\newblock \href {https://doi.org/10.18653/v1/N18-1169} {Delete, retrieve,
  generate: a simple approach to sentiment and style transfer}.
\newblock In \emph{Proceedings of the 2018 Conference of the North {A}merican
  Chapter of the Association for Computational Linguistics: Human Language
  Technologies, Volume 1 (Long Papers)}, pages 1865--1874, New Orleans,
  Louisiana. Association for Computational Linguistics.

\bibitem[{Li et~al.(2019)Li, Kiseleva, and de~Rijke}]{li2019dialogue}
Ziming Li, Julia Kiseleva, and Maarten de~Rijke. 2019.
\newblock \href {https://arxiv.org/pdf/1812.03509.pdf} {Dialogue generation:
  From imitation learning to inverse reinforcement learning}.
\newblock In \emph{Proceedings of the AAAI Conference on Artificial
  Intelligence}, volume~33, pages 6722--6729.

\bibitem[{Liu et~al.(2021)Liu, Jia, and Vosoughi}]{liu2021transformer}
Ruibo Liu, Chenyan Jia, and Soroush Vosoughi. 2021.
\newblock A transformer-based framework for neutralizing and reversing the
  political polarity of news articles.
\newblock \emph{Proceedings of the ACM on Human-Computer Interaction},
  5(CSCW1):1--26.

\bibitem[{Liu et~al.(2019)Liu, Ott, Goyal, Du, Joshi, Chen, Levy, Lewis,
  Zettlemoyer, and Stoyanov}]{liu2019roberta}
Yinhan Liu, Myle Ott, Naman Goyal, Jingfei Du, Mandar Joshi, Danqi Chen, Omer
  Levy, Mike Lewis, Luke Zettlemoyer, and Veselin Stoyanov. 2019.
\newblock Roberta: A robustly optimized bert pretraining approach.
\newblock \emph{arXiv preprint arXiv:1907.11692}.

\bibitem[{Miyato and Koyama(2018)}]{miyato2018cgans}
Takeru Miyato and Masanori Koyama. 2018.
\newblock \href {http://xxx.itp.ac.cn/pdf/2001.11314.pdf} {cgans with
  projection discriminator}.
\newblock In \emph{International Conference on Learning Representations}.

\bibitem[{Park et~al.(2018)Park, Cho, and Kim}]{park2018hierarchical}
Yookoon Park, Jaemin Cho, and Gunhee Kim. 2018.
\newblock \href {https://www.aclweb.org/anthology/N18-1162.pdf} {A hierarchical
  latent structure for variational conversation modeling}.
\newblock In \emph{Proceedings of the 2018 Conference of the North American
  Chapter of the Association for Computational Linguistics: Human Language
  Technologies, Volume 1 (Long Papers)}, pages 1792--1801.

\bibitem[{Radford et~al.(2019)Radford, Wu, Child, Luan, Amodei, and
  Sutskever}]{radford2019language}
Alec Radford, Jeffrey Wu, Rewon Child, David Luan, Dario Amodei, and Ilya
  Sutskever. 2019.
\newblock \href
  {https://cdn.openai.com/better-language-models/language_models_are_unsupervised_multitask_learners.pdf}
  {Language models are unsupervised multitask learners}.
\newblock \emph{OpenAI Blog}, 1(8):9.

\bibitem[{Raffel et~al.(2020)Raffel, Shazeer, Roberts, Lee, Narang, Matena,
  Zhou, Li, and Liu}]{raffel2020exploring}
Colin Raffel, Noam Shazeer, Adam Roberts, Katherine Lee, Sharan Narang, Michael
  Matena, Yanqi Zhou, Wei Li, and Peter~J Liu. 2020.
\newblock Exploring the limits of transfer learning with a unified text-to-text
  transformer.
\newblock \emph{Journal of Machine Learning Research}, 21:1--67.

\bibitem[{Ramachandran et~al.(2017)Ramachandran, Zoph, and
  Le}]{ramachandran2017searching}
Prajit Ramachandran, Barret Zoph, and Quoc~V Le. 2017.
\newblock \href {https://arxiv.org/pdf/1710.05941.pdf} {Searching for
  activation functions}.
\newblock \emph{arXiv preprint arXiv:1710.05941}.

\bibitem[{Rothe et~al.(2020)Rothe, Narayan, and Severyn}]{rothe2020leveraging}
Sascha Rothe, Shashi Narayan, and Aliaksei Severyn. 2020.
\newblock \href {https://www.aclweb.org/anthology/2020.tacl-1.18.pdf}
  {Leveraging pre-trained checkpoints for sequence generation tasks}.
\newblock \emph{Transactions of the Association for Computational Linguistics},
  8:264--280.

\bibitem[{Rush et~al.(2015)Rush, Chopra, and Weston}]{rush2015neural}
Alexander~M Rush, Sumit Chopra, and Jason Weston. 2015.
\newblock \href {https://www.aclweb.org/anthology/D15-1044.pdf} {A neural
  attention model for abstractive sentence summarization}.
\newblock In \emph{Proceedings of the 2015 Conference on Empirical Methods in
  Natural Language Processing}, pages 379--389.

\bibitem[{Salovey and Sluyter(1997)}]{salovey1997emotional}
Peter~Ed Salovey and David~J Sluyter. 1997.
\newblock \emph{Emotional development and emotional intelligence: Educational
  implications.}
\newblock Basic Books.

\bibitem[{Schmidt(2019)}]{schmidt-2019-generalization}
Florian Schmidt. 2019.
\newblock \href {https://doi.org/10.18653/v1/D19-5616} {Generalization in
  generation: A closer look at exposure bias}.
\newblock In \emph{Proceedings of the 3rd Workshop on Neural Generation and
  Translation}, pages 157--167, Hong Kong. Association for Computational
  Linguistics.

\bibitem[{Seo et~al.(2017)Seo, Kembhavi, Farhadi, and
  Hajishirzi}]{seo2016bidirectional}
Minjoon Seo, Aniruddha Kembhavi, Ali Farhadi, and Hannaneh Hajishirzi. 2017.
\newblock \href {https://arxiv.org/pdf/1611.01603.pdf} {Bidirectional attention
  flow for machine comprehension}.
\newblock In \emph{5th International Conference on Learning Representations,
  ICLR 2017}.

\bibitem[{Serban et~al.(2016)Serban, Sordoni, Bengio, Courville, and
  Pineau}]{serban2016building}
Iulian~V Serban, Alessandro Sordoni, Yoshua Bengio, Aaron Courville, and Joelle
  Pineau. 2016.
\newblock \href
  {https://www.aaai.org/ocs/index.php/AAAI/AAAI16/paper/view/11957/12160}
  {Building end-to-end dialogue systems using generative hierarchical neural
  network models}.

\bibitem[{Serban et~al.(2017)Serban, Sordoni, Lowe, Charlin, Pineau, Courville,
  and Bengio}]{serban2017hierarchical}
Iulian~Vlad Serban, Alessandro Sordoni, Ryan Lowe, Laurent Charlin, Joelle
  Pineau, Aaron Courville, and Yoshua Bengio. 2017.
\newblock \href
  {https://www.aaai.org/ocs/index.php/AAAI/AAAI17/paper/view/14567/14219} {A
  hierarchical latent variable encoder-decoder model for generating dialogues}.
\newblock In \emph{Thirty-First AAAI Conference on Artificial Intelligence}.

\bibitem[{Shen and Feng(2020)}]{shen-feng-2020-cdl}
Lei Shen and Yang Feng. 2020.
\newblock \href {https://doi.org/10.18653/v1/2020.acl-main.52} {{CDL}:
  Curriculum dual learning for emotion-controllable response generation}.
\newblock In \emph{Proceedings of the 58th Annual Meeting of the Association
  for Computational Linguistics}, pages 556--566, Online. Association for
  Computational Linguistics.

\bibitem[{Shen et~al.(2017)Shen, Su, Li, Li, Niu, Zhao, Aizawa, and
  Long}]{shen-etal-2017-conditional}
Xiaoyu Shen, Hui Su, Yanran Li, Wenjie Li, Shuzi Niu, Yang Zhao, Akiko Aizawa,
  and Guoping Long. 2017.
\newblock \href {https://doi.org/10.18653/v1/P17-2080} {A conditional
  variational framework for dialog generation}.
\newblock In \emph{Proceedings of the 55th Annual Meeting of the Association
  for Computational Linguistics (Volume 2: Short Papers)}, pages 504--509,
  Vancouver, Canada. Association for Computational Linguistics.

\bibitem[{Song et~al.(2019)Song, Zheng, Liu, Xu, and
  Huang}]{song-etal-2019-generating}
Zhenqiao Song, Xiaoqing Zheng, Lu~Liu, Mu~Xu, and Xuanjing Huang. 2019.
\newblock \href {https://doi.org/10.18653/v1/P19-1359} {Generating responses
  with a specific emotion in dialog}.
\newblock In \emph{Proceedings of the 57th Annual Meeting of the Association
  for Computational Linguistics}, pages 3685--3695, Florence, Italy.
  Association for Computational Linguistics.

\bibitem[{Tay et~al.(2021)Tay, Bahri, Metzler, Juan, Zhao, and
  Zheng}]{tay2020synthesizer}
Yi~Tay, Dara Bahri, Donald Metzler, Da-Cheng Juan, Zhe Zhao, and Che Zheng.
  2021.
\newblock \href {https://arxiv.org/pdf/2005.00743.pdf} {Synthesizer: Rethinking
  self-attention in transformer models}.
\newblock In \emph{Proceedings of the 37th International Conference on Machine
  Learning (ICML 2021)}.

\bibitem[{Tolstikhin et~al.(2021)Tolstikhin, Houlsby, Kolesnikov, Beyer, Zhai,
  Unterthiner, Yung, Keysers, Uszkoreit, Lucic et~al.}]{tolstikhin2021mlp}
Ilya Tolstikhin, Neil Houlsby, Alexander Kolesnikov, Lucas Beyer, Xiaohua Zhai,
  Thomas Unterthiner, Jessica Yung, Daniel Keysers, Jakob Uszkoreit, Mario
  Lucic, et~al. 2021.
\newblock \href {https://arxiv.org/pdf/2105.01601.pdf} {Mlp-mixer: An all-mlp
  architecture for vision}.

\bibitem[{Vaswani et~al.(2017)Vaswani, Shazeer, Parmar, Uszkoreit, Jones,
  Gomez, Kaiser, and Polosukhin}]{NIPS2017_7181}
Ashish Vaswani, Noam Shazeer, Niki Parmar, Jakob Uszkoreit, Llion Jones,
  Aidan~N Gomez, \L~ukasz Kaiser, and Illia Polosukhin. 2017.
\newblock \href
  {http://papers.nips.cc/paper/7181-attention-is-all-you-need.pdf} {Attention
  is all you need}.
\newblock In I.~Guyon, U.~V. Luxburg, S.~Bengio, H.~Wallach, R.~Fergus,
  S.~Vishwanathan, and R.~Garnett, editors, \emph{Advances in Neural
  Information Processing Systems 30}, pages 5998--6008. Curran Associates, Inc.

\bibitem[{Wei et~al.(2019)Wei, Liu, Mao, Guo, Zhu, Zhou, and
  Hu}]{wei2019emotion}
Wei Wei, Jiayi Liu, Xianling Mao, Guibing Guo, Feida Zhu, Pan Zhou, and Yuchong
  Hu. 2019.
\newblock \href {https://dl.acm.org/doi/pdf/10.1145/3357384.3357937}
  {Emotion-aware chat machine: Automatic emotional response generation for
  human-like emotional interaction}.
\newblock In \emph{Proceedings of the 28th ACM International Conference on
  Information and Knowledge Management}, pages 1401--1410.

\bibitem[{Xu et~al.(2015)Xu, Wang, Chen, and Li}]{xu2015empirical}
Bing Xu, Naiyan Wang, Tianqi Chen, and Mu~Li. 2015.
\newblock \href {https://arxiv.org/pdf/1505.00853.pdf} {Empirical evaluation of
  rectified activations in convolutional network}.
\newblock \emph{arXiv preprint arXiv:1505.00853}.

\bibitem[{Yang et~al.(2020)Yang, Wu, Xu, Liang, Bai, Wang, Wang, and
  Li}]{yang-etal-2020-styledgpt}
Ze~Yang, Wei Wu, Can Xu, Xinnian Liang, Jiaqi Bai, Liran Wang, Wei Wang, and
  Zhoujun Li. 2020.
\newblock \href {https://doi.org/10.18653/v1/2020.findings-emnlp.140}
  {{S}tyle{DGPT}: Stylized response generation with pre-trained language
  models}.
\newblock In \emph{Findings of the Association for Computational Linguistics:
  EMNLP 2020}, pages 1548--1559, Online. Association for Computational
  Linguistics.

\bibitem[{Yang et~al.(2019{\natexlab{a}})Yang, Wu, Yang, Xu, and
  Li}]{yang-etal-2019-low}
Ze~Yang, Wei Wu, Jian Yang, Can Xu, and Zhoujun Li. 2019{\natexlab{a}}.
\newblock \href {https://doi.org/10.18653/v1/D19-1197} {Low-resource response
  generation with template prior}.
\newblock In \emph{Proceedings of the 2019 Conference on Empirical Methods in
  Natural Language Processing and the 9th International Joint Conference on
  Natural Language Processing (EMNLP-IJCNLP)}, pages 1886--1897, Hong Kong,
  China. Association for Computational Linguistics.

\bibitem[{Yang et~al.(2019{\natexlab{b}})Yang, Dai, Yang, Carbonell,
  Salakhutdinov, and Le}]{yang2019xlnet}
Zhilin Yang, Zihang Dai, Yiming Yang, Jaime Carbonell, Russ~R Salakhutdinov,
  and Quoc~V Le. 2019{\natexlab{b}}.
\newblock \href
  {http://papers.nips.cc/paper/8812-xlnet-generalized-autoregressive-pretraining-for-language-understanding.pdf}
  {Xlnet: Generalized autoregressive pretraining for language understanding}.
\newblock In \emph{Advances in neural information processing systems}, pages
  5754--5764.

\bibitem[{Yu et~al.(2017)Yu, Zhang, Wang, and Yu}]{yu2017seqgan}
Lantao Yu, Weinan Zhang, Jun Wang, and Yong Yu. 2017.
\newblock \href
  {https://www.aaai.org/ocs/index.php/AAAI/AAAI17/paper/view/14344/14489}
  {Seqgan: Sequence generative adversarial nets with policy gradient}.
\newblock In \emph{Thirty-first AAAI conference on artificial intelligence}.

\bibitem[{Zhang et~al.(2019{\natexlab{a}})Zhang, Goodfellow, Metaxas, and
  Odena}]{zhang2019self}
Han Zhang, Ian Goodfellow, Dimitris Metaxas, and Augustus Odena.
  2019{\natexlab{a}}.
\newblock \href {https://arxiv.org/pdf/1805.08318.pdf} {Self-attention
  generative adversarial networks}.
\newblock In \emph{International Conference on Machine Learning}, pages
  7354--7363.

\bibitem[{Zhang et~al.(2019{\natexlab{b}})Zhang, Kishore, Wu, Weinberger, and
  Artzi}]{bertscore}
Tianyi Zhang, Varsha Kishore, Felix Wu, Kilian~Q. Weinberger, and Yoav Artzi.
  2019{\natexlab{b}}.
\newblock \href {http://arxiv.org/abs/1904.09675} {Bertscore: Evaluating text
  generation with {BERT}}.
\newblock \emph{CoRR}, abs/1904.09675.

\bibitem[{Zhang et~al.(2019{\natexlab{c}})Zhang, Feng, Meng, You, and
  Liu}]{zhang-etal-2019-bridging}
Wen Zhang, Yang Feng, Fandong Meng, Di~You, and Qun Liu. 2019{\natexlab{c}}.
\newblock \href {https://doi.org/10.18653/v1/P19-1426} {Bridging the gap
  between training and inference for neural machine translation}.
\newblock In \emph{Proceedings of the 57th Annual Meeting of the Association
  for Computational Linguistics}, pages 4334--4343, Florence, Italy.
  Association for Computational Linguistics.

\bibitem[{Zhang et~al.(2020)Zhang, Sun, Galley, Chen, Brockett, Gao, Gao, Liu,
  and Dolan}]{zhang-etal-2020-dialogpt}
Yizhe Zhang, Siqi Sun, Michel Galley, Yen-Chun Chen, Chris Brockett, Xiang Gao,
  Jianfeng Gao, Jingjing Liu, and Bill Dolan. 2020.
\newblock \href {https://doi.org/10.18653/v1/2020.acl-demos.30} {{DIALOGPT} :
  Large-scale generative pre-training for conversational response generation}.
\newblock In \emph{Proceedings of the 58th Annual Meeting of the Association
  for Computational Linguistics: System Demonstrations}, pages 270--278,
  Online. Association for Computational Linguistics.

\bibitem[{Zhao et~al.(2017)Zhao, Zhao, and Eskenazi}]{zhao-etal-2017-learning}
Tiancheng Zhao, Ran Zhao, and Maxine Eskenazi. 2017.
\newblock \href {https://doi.org/10.18653/v1/P17-1061} {Learning
  discourse-level diversity for neural dialog models using conditional
  variational autoencoders}.
\newblock In \emph{Proceedings of the 55th Annual Meeting of the Association
  for Computational Linguistics (Volume 1: Long Papers)}, pages 654--664,
  Vancouver, Canada. Association for Computational Linguistics.

\bibitem[{Zhou et~al.(2018)Zhou, Huang, Zhang, Zhu, and
  Liu}]{zhou2018emotional}
Hao Zhou, Minlie Huang, Tianyang Zhang, Xiaoyan Zhu, and Bing Liu. 2018.
\newblock \href
  {http://coai.cs.tsinghua.edu.cn/hml/media/files/aaai2018-ecm.pdf} {Emotional
  chatting machine: Emotional conversation generation with internal and
  external memory}.
\newblock In \emph{Thirty-Second AAAI Conference on Artificial Intelligence}.

\bibitem[{Zhou and Wang(2018)}]{zhou2018mojitalk}
Xianda Zhou and William~Yang Wang. 2018.
\newblock \href {https://www.aclweb.org/anthology/P18-1104.pdf} {Mojitalk:
  Generating emotional responses at scale}.
\newblock In \emph{Proceedings of the 56th Annual Meeting of the Association
  for Computational Linguistics (Volume 1: Long Papers)}, pages 1128--1137.

\end{thebibliography}


\end{document}